# Chaotic Fitness Dependent Optimizer for Planning and Engineering Design

Hardi M. Mohammed[1,2], Tarik A. Rashid[3]

[1]Applied Computer Department, College of Medicals and Applied Sciences, Charmo University, Sulaimani, Chamchamal, KRG, Iraq.

[2]Technical College of Informatics, Sulaimani Polytechnic University, Sulaimani, KRG, Iraq

[3]Computer Science and Engineering, University of Kurdistan Hewler (UKH), Erbil, KRG, Iraq.

Corresponding author: Hardi M. Mohammed (e-mail: hardi.mohammed@charmouniversity.org).

ORCID: 0000-0002-9766-9100

**ABSTRACT** Fitness Dependent Optimizer (FDO) is a recent metaheuristic algorithm that mimics the reproduction behavior of the bee swarm in finding better hives. This algorithm is similar to Particle Swarm Optimization (PSO) but it works differently. The algorithm is very powerful and has better results compared to other common metaheuristic algorithms. This paper aims at improving the performance of FDO, thus, the chaotic theory is used inside FDO to propose Chaotic FDO (CFDO). Ten chaotic maps are used in the CFDO to consider which of them are performing well to avoid local optima and finding global optima. New technic is used to conduct population in specific limitation since FDO technic has a problem to amend population. The proposed CFDO is evaluated by using 10 benchmark functions from CEC2019. Finally, the results show that the ability of CFDO is improved. Singer map has a great impact on improving CFDO while the Tent map is the worst. Results show that CFDO is superior to GA, FDO, and CSO. Both CEC2013 and CEC2005 are used to evaluate CFDO. Finally, the proposed CFDO is applied to classical engineering problems, such as pressure vessel design and the result shows that CFDO can handle the problem better than WOA, GWO, FDO, and CGWO. Besides, CFDO is applied to solve the task assignment problem and then compared to the original FDO. The results prove that CFDO has better capability to solve the problem.

## 1. INTRODUCTION

Finding an optimal solution is the aim of solving optimization problems while they have various constraints. The optimal solution should be found in a feasible search space and in a reasonable amount of time. Usually, traditional algorithms have been used to solve optimization problems. These algorithms require a lot of time to achieve an optimal solution because they are searching from a single point and attempting to collect much information [1]. Traditional algorithms, such as Hill climbing, Simulated Annealing (SA), and Random Search, and then, metaheuristic algorithms have been used to solve complex optimization problems [2]. However, these algorithms also have some deficiencies in terms of finding global optima due to having different constraints in real-world applications, namely, engineering design problems [3], task planning problems, and economical problems [4].

Metaheuristic algorithms are used to solve real-world problems either by taking physical or biological phenomenon [5], [6]. Depending on the stochastic technics, metaheuristic algorithms could avoid local optima and obtain global optima while they have different values and convergence speeds [7], [8]. Recent metaheuristic algorithms, such as Particle Swarm Optimization (PSO) [9], Bat Optimization Algorithm (BA) [10], Grey Wolf Optimization Algorithm (GWO) [11], and Whale Optimization Algorithm (WOA) [12] are popular because of two factors: Firstly, they are very simple, efficient, robust, and adaptable. Secondly, in terms of implementation, they can be implemented easily in any programming language. Because of these two factors, they are used to solve various real-world problems [13], [14], [15].

Even though finding local optima and global optima depend on the search mechanism inside the algorithm, furthermore, balancing between exploration and exploitation is another crucial point in metaheuristic algorithms because it depends on stochastic numbers. Achieving better performance can be done by having a proper balance between exploration and exploitation [13], [16]. Exploitation uses information that has been collected so far to achieve a solution better than the previous one in the next generation [17], [18]. In contrast, exploration helps the algorithm avoid local optima and obtaining a new solution that can be far from the current solution [19], [20]. Therefore, avoiding local optima is the advantage of exploration while having slow convergence is the drawback of it. However, the advantage of exploitation is high convergence while it may lead to stuck in local optima [13], [21].

FDO is a nature metaheuristic algorithm that is based on the reproductive behavior of bee swarms while the bees are trying for finding new hives. FDO is based on PSO but it uses a different mechanism to update agent position. Scout bees search for positions randomly, while the better position is found, the previous position is neglected. If the next position is not better than the previous position, then the FDO uses the previous position to find a better solution [22]. FDO has a very competitive result compared to other metaheuristic algorithms and it has also been used to solve some real-world problems, such as frequency-modulated sound waves, aperiodic antenna array designs, pedestrian evacuation models, etc. [10] [23].

Trapping into the local optima is a vital matter due to the random initialization of variables and population. Hence, to circumvent the above problem chaos theory is used in FDO. Consequently, in this paper, FDO and CFDO are applied to the classical engineering problem, which is called pressure vessel design.

There are three ways to use chaos theory inside metaheuristic algorithms. Firstly, chaos can be used to initialize random numbers in the algorithm. Using chaos as the operation is the second way of using chaos, and finally, it can be used to initialize the population [4], [24]. Using these techniques can improve algorithms in terms of convergence speed and avoiding local optima. Chaos theory has been used by many metaheuristic algorithms, such as  PSO [25], Artificial Bee Colony (ABC) [26], Bat Algorithm (BA) [14], Grey Wolf Optimization (GWO) [16], and Whale Optimization Algorithm (WOA) [1]. Therefore, the performance of these algorithms is improved by using chaos theory. Chaos theory is selected to improve FDO because chaotic maps have unpredictable dynamic behavior which improves the performance of metaheuristic algorithms to jump out from local optima [27].

The main contribution of this paper is proposing a Chaotic Fitness Dependent Optimizer and then using it to enhance the performance of FDO and to solve the pressure vessel design problem and task assignment problem. This contribution is done by using ten chaotic maps. The following are the main parts of the contribution in FDO:

1) Initializing population using chaotic map
2) Updating speed of current bee by multiplying its position with a chaotic random value.
3) Changing amending mechanism of position if it is beyond the search space. When the new position is calculated. A mechanism is used to check whether the new position is inside the search space or not.

Both of the following points are used at the same time to improve the FDO so that it: 1) avoids falling into local optima and 2) enhances its search capability. CEC2019 benchmark test functions are used to evaluate the result of CFDO compared to the standard FDO and also Wilcoxon Rank Sum test is used to evaluate the result statistically. CEC2013 and CEC2005 benchmark functions are also used to evaluate the performance of CFDO.

The rest of the sections are organized as follows: FDO is described in detail in Section 2, the next, chaotic maps are discussed in Section 3. Then, in Section 4, CFDO is described. In section 5, the results are presented and evaluated. Then, solving the pressure vessel design problem and task assignment problem are presented in Section 6. In the final section, the conclusion and future work are presented.

## 2. FDO

FDO is proposed by Jaza M. Abdullah and Tarik A. Rashid in 2019. This algorithm is based on the searching mechanism of the bee swarm and the reproduction process of finding better hives. Scout bees are responsible for searching among many potential hives and then they decide on which hive is better than the rest.

FDO starts by initializing the scout bee population randomly inside the search space. Discovering the hive depends on the position and fitness function of each scout bee. Finding the better hive (new solution) is the main purpose of the scout bee. If the next solution is better than the previous solution, then the scout bee ignores the previous one. However, if a better solution cannot be found, then the scout bee relies on the previous solution to change its position. Scout bee can be represented as $X_i$ (i=1, 2,...,n).

Fitness weight and random walk methods are used by scout bees to search inside the search space randomly. As can be seen in Equation (1), the scout bee movement depends on the pace of changing the position of the current scout. By adding the pace, the scout tries to explore a new solution that is better than the previous one.

$$X_{i,t+1} = X_{i,t} + pace \qquad (1)$$

Where *X* represents the artificial scout bee, *i* denotes the current scout bee (search agent) and *t* is the current iteration. the movement of the scout that can be detected is represented by the *pace*, while it depends on the fitness weight *fw*. Although, randomization is the base of the *pace*. Therefore, *fw* equation can be represented as follow:

$$fw = \left|\frac{X^*_{i,t\ fitness}}{X_{i,t\ fitness}}\right| - wf \qquad (2)$$

$X^*_{i,t}$ *fitness* represents the best solution that was found so far by scout bee and $X_{i,t}$ *fitness* is the best solution of the current scout bee. The weight factor is represented by *wf* and it has the range value in [0,1]; FDO has *r* variable, which is a random number in the range of [-1, 1], this random number is initialized by using the Levy flight mechanism. Therefore, FDO has simplicity in its calculation for finding objective functions because each agent requires fitness weight and a random number to be calculated.

After initializing the scout bee randomly on the search space by using the upper and lower boundaries. Then, a global solution is obtained. Next, *fw* is calculated, which is depending on the following conditions through the below equations:

If *fw* value is equal to 1, then the *pace* can be calculated by Equation (3) if one of the three conditions is true: *fw=0 fw=1* or $X_{i,t\ fitness} = 0$.

$$pace = X_{i,t} * random \qquad (3)$$

When *fw* is greater than zero and less than 1, then the random number *r* is generated in the range [1, -1]. After that, if *r* is less than zero Equation (4) is used to calculate the *pace*. Otherwise, the *pace* is found by Equation (5).

$$pace = distance_{best_{bee}} * random \qquad (4)$$

$$pace = distance_{best_{bee}} * fw \qquad (5)$$

Where *distance_best_bee* can be calculated by differentiating the current scout bee from the best scout bee as can be seen from Equation (6).

$$distance_{best_{bee}} = X^* - X_{i,t} \qquad (6)$$

## 3. Chaotic Maps

To improve the performance of metaheuristic algorithms, different mathematical approaches are used. One of these approaches is the chaos theory, which is a nonlinear system [28], which has dynamic behaviour. Because of its effectiveness, it has been used in different areas by researchers: research optimization, synchronization [29], and chaos control [30].

Chaotic maps are bounded by a nonlinear system that has random properties. These maps are very sensitive to initial values [31]. The chaos theory was discovered by Lorenz when he found that a little change in the initialization makes a radical change in results [24]. Furthermore, chaotic maps can be initialized by choosing any value between 0 and 1; consequently, 0.7 was used in this study as the initial point of the chaotic maps [1]. Therefore, ten common chaotic maps were used for the initialization of random variables and the population of the scout bee with FDO. Table 1 shows ten chaotic maps.

Chaotic maps are described with different NIAs in this section. Researchers have used chaotic maps to improve the performance of many algorithms. For example, chaotic GA (CGA) was proposed to overcome premature convergence problems to maximize the number of iterations and reach the global optima. Two chaotic maps are used to improve the performance of GA: tent map and logistic map. These maps were used to generate random numbers in GA. The results from five benchmark functions proved that the diversity of population from CGA can jump out of local optima and it was better than the original GA and PSO. CGA also improved GA in terms of reducing the number of iterations required to solve problems [32].

Chaotic Grasshopper Optimization Algorithm (CGOA) was proposed to consider the convergence speed problem. Ten chaotic maps were used to verify which of them has higher performance with GOA. The results revealed that the circle map had better results when it was tested on 13 benchmark functions. The aim of using a chaotic map was to improve a balance between the exploration and exploitation phases. The results showed that CGOA obtained better results by using proper balance and avoiding local optima against GOA [33].

The investigation into a proper balance between exploration and exploitation challenges researchers because metaheuristic algorithms are based on stochastic behaviour. Thus, the exploration phase uses random behaviour to find better search space, while achieving solutions in a fast way is the aim in the exploitation phase. Therefore, a chaotic GSA was proposed to overcome the problems. Adaptive normalization was also used inside GSA to transit between the exploration phase and the exploitation phase. Then, 12 benchmark functions were used to test the performance of GSA. The statistical results showed that the sinusoidal map achieved significant results compared with the original GSA [34].

Fruit fly optimization algorithm (FOA) was combined with chaos methods to overcome stagnating into local optima, and the proposed algorithm was called chaotic FOA (CFOA). Therefore, a logistic chaotic map was used to investigate the optimum best fruit fly, and the new location was based on the best location of the fruit fly and the best chaotic position. Balancing between both phases depended on a chaotic map or fruit fly independently. The simulation results based on testing six benchmark functions showed that CFOA achieved better solutions against PSO, ABC, FOA [35].

CS has difficulties in terms of convergence speed and solution quality because it is based on a random walk with a scaling factor. Chaotic CS (CCS) was proposed to avoid local optima and improve the solution quality. Consequently, the chaotic map used a variable for both scaling factor and fractional probabilities. Controlling dimension expectation was based on the fraction probability. Therefore, 20 benchmark test functions were used to test CCS to ensure that it works well when it dealt with the high dimensionality of problems [36].

According to the study used seven chaotic maps to improve ABC because ABC has problems relating to local optima and convergence speed. Chaotic maps were used to generate a random number that was required by the original ABC. Three benchmark functions were used to verify the performance of CABC. The results indicated that CABC performs well against the original ABC [37].

## 4. Chaotic Fitness Dependent Optimizer (CFDO)

Original FDO has achieved better performance compared to DA, WOA, and SSA when tested on ten CEC2019 benchmark test functions. By looking at the results of FDO in [22], it can be said that FDO still requires enhancement due to avoiding local optima and increase the convergence rate. In general, chaos can be called a source of randomness. Chaotic derives from a chaotic map, thus, chaos is the feature of the complex system and the map is the relation of the chaos behavior by using some parameters inside the algorithm [14]. Consequently, chaotic maps have been used by researchers in the field of optimization algorithms because of their nonlinear and dynamic behavior [16].

Therefore, to introduce CFDO, 10 chaotic maps are used in FDO while each of them has a different mathematical equation. As a result, CFDO is introduced in this paper to tackle this problem and improve the performance of FDO concerning the convergence speed and local optima. The pseudocode of CFDO is illustrated in Algorithm 1.

Chaotic maps are embedded in FDO in two different ways:

1) Initialization of the random variable *r*, which is generated by using a chaotic map instead of using the Levy flight method. This variable is used to update *pace* by multiplying position. New technic is also used to amend the population if they are outside the limitation.

2) Initializing the scout bee population using a chaotic map instead of the stochastic mechanism.

**Algorithm 1 Pseudocode of the CFDO**

**Initialize the population of scout bee by using chaotic map $X_{i,t}$ (i=1,2,…, n)**
**while** (t< max_iteration)
    **for** each scout bee $X_{i,t}$
        Select best scout bee $X^*_{i,t}$

        **If** ($X_{i,t}$ fitness==0)
            *fitness weight* =0
        **else**
        find *fitness weight* using Equation (2)
        **end if**
        **if** (fitness weight=0 // fitness weight=1 // $X_{i,t}$ fitness==0)
            find *pace* by using Equation *(3)*
        **else**
            **r generated in range [-1, 1] by using chaotic map**
            **if** (r>=0)
              calculate the *pace* using Equation (5)
            **else**
              calculate *pace* using Equation (4)
            **end if**
        **end if**
        determine $X_{i,\,t+1}$ using Equation (1)
        **if** ($X_{i,\,t+1}$< $X_{i,t}$ fitness)
        update position of scout and select *pace*
        **else**

```
                    calculate X_{i, t+1} using Equation (1)
                        if (X_{i, t+1} < X_{i,t} fitness)
                        update position of scout and select pace
                        else
                        update current position without any movement
                        end if
                    end if
                end for
end while
```

## 5. EXPERIMENTAL RESULTS AND DISCUSSION

According to [1], [14] ten chaotic maps have been used to initialize a random variable and population of scout bee in the proposed chaotic algorithms and then compared against the FDO. As a result, the ten chaotic maps are used separately inside the FDO for both initializing the random value and initializing the scout bee population.

Therefore, to ensure that the performance of any optimization algorithms is efficient, they must be tested by common benchmark test functions. As a result, different benchmark functions, such as CEC2005, CEC2013, and CEC2019 are used to evaluate the proposed CFDO.

### 5.1 CEC2019 Benchmark Functions

Parameters are set as 30 is taken as a population size for CFDO, GA, PSO, CSO, FDO. Each algorithm performed 100 iterations to find the optimum solution. Then, ten chaotic maps are used inside CFDO. As a result, the Singer map performed superior compared to other maps. This algorithm runs 30 times and then the average results of the 30 times are taken for CEC2019 benchmark functions.

As a result, CEC2019 benchmark functions are proposed in [38] and used in [39], and then, the implementation is conducted using Matlab code to verify the performance of our proposed chaotic FDO. CEC2019 consists of ten benchmark functions. All the functions in CEC2019 are multimodal. Therefore, multimodal functions have more than one optimum value, accordingly, it is a challenging matter for the CFDO, because it has to find one global optimum and avoid others, which is called local optima. Table 2 presents CEC2019 functions, which show the functions name, dimension, and the range of the search space for each function.

#### 5.1.1 Comparing CFDO Vs. FDO using different chaotic maps

Results are conducted from various chaotic maps by taking 50 iterations and 30 population sizes. The average results are taken from 30 independent runs. Ten chaotic maps are applied to the DFO, namely, Chebyshev, Circle, Iterative, Logistic, Piecewise, Sine, Singer, Sinusoidal, and Tent. These maps are presented in Table 1 as CFDO1 to CFDO 10 respectively.

Tables 3, 4, 5, 6, and 7 show the results of each chaotic map against FDO. Each table includes the results of two chaotic maps. Table 5 shows that CFDO5 has superior results compared to FDO and other chaotic FDOs. CFDO 6 and CFDO 7 have the second and third best averages respectively. Therefore, CFDO 1 to CFDO 9 has improved the performance of FDO. In contrast, Table 7 illustrates that CFDO 10 has produced the worst result against FDO. It can be said that Chebyshev, Circle, Iterative, Logistic, Piecewise, Sine, Singer, and Sinusoidal could improve the performance of FDO. The $p$-value showed that CFDO 8 has six significant values and it is the best one and followed by CFDO 5 and CFDO 6, which their results are significant in 5 benchmark functions out of ten. Overall, the Tent map shows the worse results in all ten benchmark functions. However, the Singer map enhances the performance of CFDO and recorded the first best result.

### 5.1.2 Statistical Results

Evaluating performance based on the average and standard deviation is not accurate enough, hence, the statistical test must be used. Depends on the statistical test, the proposed chaotic algorithm can be evaluated against other metaheuristic algorithms. Therefore, Wilcoxon rand-sum test is conducted in this paper to evaluate the mean and standard deviation to ensure that either the result is significant or not. In general, the $p$-value is calculated using the Wilcoxon rank-sum test and must be less than 0.05, then the result is significant. Tables 3, 4, 5, 6, and 7 present the $p$-value for the results, and the significant results are underlined.

It can be seen that the Singer map shows the best significant results in 6 out of 10 functions while the Circle map and Tent map provide the worst results. Furthermore, the Logistic map and Piecewise map have the second-best significant value in the 5 functions. Sine map also provides significant value in 4 functions while Gauss/ Mouse, Iterative and Sinusoidal map has 3 significant values. However, the second-worst result is the Circle map, which has only one single significant result.

Overall, Singer, Logistic, and Piecewise maps are better chaotic maps that could enhance the performance of CFDO. The reason behind this improvement is that these three chaotic maps can improve the exploration capability of CFDO and avoid local optima and also find the best global optima among various solutions.

As it is clear that from Table 8 Singer map is the best chaotic map that improves the performance of FDO. Consequently, CFDO is evaluated on the CEC2019 benchmark functions and then compared to CSO, GA, PSO, and FDO. It can be seen that from Table 8 that CFDO has outperformed FDO in six functions. However, CFDO and FDO have the same result in function three. CFDO is also better than GA and CSO in six functions while it outperforms well in one function against PSO. Overall, CFDO performance is better than FDO, GA, and CSO. However, PSO achieved the best results compared to CFDO.

The reason behind these results is that using the chaotic singer map for the initializing population and the random number is very efficient. Finally, results show that the proposed CFDO is very efficient and can be applied to solve real-world problems.

## 5.2 EC2013 Benchmark Functions

The population size is set to 100 search agents with 100 dimensions for each test function. The algorithm runs 5000 iterations for each time. The results of the compared algorithms are taken from 51 run times. Then, CFDO is compared against the algorithms.

Another common benchmark function that is used to evaluate CFDO is the CEC2013 benchmark function. CEC2013 consists of 28 benchmark functions which are: F1-F5 unimodal functions, F6-F20 multimodal functions, and F21-F28 composite functions. Table 9 presents the benchmark functions of CEC2013.

CFDO is compared with the different common algorithms, such as Gravitation Search Algorithm (GSA), Chaotic (CJADE), Success History-based parameter Adaptation for Differential Evolution (SHADE), Increasing Population covariance matrix adaptation evolution strategy (IPOP-CMA-ES), and NBIPOP-CMA-ES. The results of the algorithms are taken from [40], [41], [42], [43].

Table 10 shows the comparison of CFDO against CJADE, IPOP-CMA-ES, NIPOP-CMA-ES, SHADE, and GSA. Results proved that CFDO has the worst results compared to the other algorithms. However, it has the same result as GSA and CJADE in function eight. Despite having poor results, CFDO achieved better results in function nine against SHADE, NBIPOP-CMA-ES, and IPOP-CMA-ES. As it can be seen that CFDO did not have better results compared to the other algorithm because of the complexity of CEC2013 benchmark functions. Also, it can be said that chaotic map techniques did not perform well to solve CEC2013 benchmark functions.

## 5.3 CEC2005 Benchmark Functions

To evaluate the performance of CFDO, 14 benchmark functions were used CFDO. The benchmark functions are unimodal, multimodal, and expanded benchmark functions. F1-F5 are unimodal functions while F6-F12 are multimodal functions. Both functions 13 and 14 are expanded benchmark functions. Table 11 shows the CEC2005 benchmark functions. The population size using CEC2005 benchmark functions is set to 100 with 30 dimensions. Each algorithm runs 30 times with 5000 iterations. The results of the comparison are taken from the average of the 30 runs. Table 12 presents that CFDO compared against BSO and CBSO. The results of BSO and CBSO were taken from [27]. As it can be seen from the table CFDO outperforms well compared to BSO and CBSO in 7 functions out of 14 functions. CFDO is better than CBSO in three unimodal functions which means that it performs well in terms of exploration. However, CBSO is superior regarding the exploitation phase because CFDO is better in only 2 functions out of 7. The reason behind the improvement of CFDO in exploration is that using a chaotic map improved the randomization of the population. Also using the chaotic technique instead of the Levy Random Number technique improved the performance of FDO because the random number is responsible for updating the *pace* value in FDO.

## 6. Solving Application Problems by CFDO

In this section, CFDO is used to solve two problems, which are pressure vessel design and task assignment problems.

### 6.1 Pressure Vessel Design Problem

Classical engineering problem like pressure vessel design is a problem. The cost of three sections of the cylindrical pressure vessel is required to be optimized, so, minimizing the cost is the aim of optimizing this problem. Forming, material, and welding should be minimized. The head of the vessel has a hemispherical shape while the end of both sides of the vessel is crapped. Four variables have to be optimized. These variables are the head thickness $T_h$, shell thickness $T_s$, inner radius $R$ and cylindrical length section without counting the head $L$. As a result, the problem has to be minimized depends on four constraints. The following equations describe the constraints of the problem.

$$n = 1,2,\ldots,4$$

$$\vec{x} = [x_1 x_2 x_3 x_4] = [T_s T_h\ R\ L],$$

$$f(\vec{x}) = 0.6224 x_1 x_3 x_4 + 1.7781 x_2 x_3^2 + 3.1661 x_1^2 x_4 + 19.84 x_1^2 x_3, \tag{7}$$

Variable limitation

$$0 \le x_1 \le 99,$$

$$0 \le x_2 \le 99,$$

$$10 \le x_3 \le 200,$$

$$10 \le x_4 \le 200,$$

These are subjected to

$$g_1(\vec{x}) = -x_1 + 0.0193 x_3 \le 0 \tag{8}$$

$$g_2(\vec{x}) = -x_3 + 0.00954x_3 \leq 0 \tag{9}$$

$$g_3(\vec{x}) = -\pi x_3^2 x_4 - \frac{4}{3}\pi x_3^3 + 1{,}296{,}000 \leq 0 \tag{7}$$

$$g_4(\vec{x}) = x_4 + 240 \leq 0 \tag{8}$$

Different algorithms have been used to solve this problem, such as GA, WOA, PSO, and GWO. Therefore, the authors applied FDO and CFDO to solve the problem. Table 13 shows that CFDO outperforms well compared to other algorithms while CGWO has the second-best result after CFDO. Finally, CFDO obtained these results 1.54, 6.10, 30.58, 73.29 for $T_s, T_h, R$ and $L$ respectively.

## 6.2 Solving Task Assignment Problem

The task assignment problem is an NP problem and it is the base step to ensure the ability of exploitation in parallel or distributed computing systems. It is working on diving tasks among different processors [44] [45]. The aim of optimizing this problem is to minimize the execution cost for each task. Task assignment problem has the following equations:

$$f(x) = \sum_{i=1}^{n}\sum_{j=1}^{n} C_{ij} X_{ij} \tag{12}$$

Subject to:

$$X_{ij} = \begin{cases} 1 \text{ if the task is assigned to specific processor,} \\ 0, \quad otherwise \end{cases}$$

Where $C_{ij}$ is the cost of each task while $X_{ij}$ is the indication for the task whether it is assigned for a specific processor or not. In this paper, the authors have task assignment problems related to a department. The problem is that we have five employees with five tasks. Each employee can execute the task at different times. The execution time for all the tasks has a lower limit and upper limit, which are 10 and 99 respectively. Table 14 illustrates the tasks and employees as an example. Therefore, CFDO and FDO are used to solve this problem to show the efficiency of the proposed CFDO compared to FDO.

CFDO and FDO algorithms have been run for 15 iterations because they could solve the problem in several iterations. As it can be seen that from Table 15, CFDO outperforms well and its performance is much better than FDO due to using chaotic maps inside FDO.

## 7. Result Analysis

To analyze the results of CFDO, few points should be taken into consideration. Initializing the population of metaheuristic algorithms is a critical point by researchers. So, different technic of randomizations is used to have proper search space. However, in this paper using the chaotic randomization technique improved the performance of FDO because chaotic maps have non-repetition and ergodicity features which help the algorithm to explore the search space at a higher speed.

The *pace* vector has a great role inside FDO to update the position to define the new solution. A random variable, which is called *r* is a key parameter that can be tuned to change the value of *pace* while the new position is updated based on the *pace* value. In addition, using a chaotic map improved the exploration of FDO because of having great dynamic randomization behavior of the chaotic maps. As a result, the main reason behind the great achievement of CFDO against the compared algorithm is using chaotic maps to initialize the population of the scout bee and initializing the random variable (*r*).

## 8. Conclusion and Future Work

In this paper chaotic theory is embedded inside the FDO algorithm, thus, the CFDO is proposed. The CFDO is presented to enhance the performance of CFDO. To make sure that the chaotic maps have a great impact on the performance of CFDO, chaotic maps were used in two different ways at the same time. The chaotic map is used to initialize the random variable and to initialize the scout bee population. Ten different chaotic maps were used in the CFDO algorithm, then, the CFDO was evaluated using CEC2019 benchmark functions, which consisted of ten benchmark functions. Wilcoxon's rand-sum test was also calculated to ensure that the results were either significant or not. In general, chaotic maps improved the ability of CFDO for searching for global optima and avoiding local optima because chaotic maps could control the random variable and population of scout bee for finding the optimum solution and improving the convergence of CFDO. The results showed that the Singer map had the best results compared to the original FDO. Although, the Tent map achieved the worst results.

CEC2019 benchmark functions were used to evaluate the performance of CFDO against FDO, GA, CSO, and PSO. Results indicated that CFDO could achieve higher performance compared to FDO, GA, and CSO. However, PSO performed well against CFDO. CFDO was used to solve CEC2013 benchmark functions and the results proved that CFDO did not achieve the best results compared to the other five algorithms. However, it obtained high performance when it was used to solve CEC2005 functions. Therefore, CFDO achieves seven best results out of 14 compared to BSO and CBSO. FDO and CFDO were applied to pressure vessel design as an engineering problem and then compared to WOA, GWO, and CGWO. Results proved that CFDO improved the ability of FDO to solve the problem. Then, the task assignment problem has been solved by CFDO and FDO. Results proved that CFDO was better at solving the problem.

Chaotic maps could improve the performance of FDO and avoid local optima because of their dynamic and superlative way of generating random numbers. Using new technic to amend the bee population if they are outside the limitation is another reason which enhanced the capacity of FDO. Since using its old technic to solve task assignment problems, FDO has problems with amending the population. Also, CFDO can be further extended to solve mixed-type problems.

## Compliance with Ethical Standards

**Conflict of Interest:** The authors declare that they have no conflict of interest.

**Ethical approval:** This article does not contain any studies with human participants or animals performed by any of the authors.

**Table 1 Chaotic Maps**

| | Chaotic map | Range |
|---|---|---|
| Chebyshev [32] | $x_{i+1} = \cos(i\, x^{-1}(x_i))$ | (-1,1) |
| Circle [33] | $x_{i+1} = \left(x_i + b - \left(\frac{a}{2\pi}\right)\sin(2\pi x_k), 1\right), a = 0.5, b = 0.2$ | (0,1) |
| Gauss/mouse [34] | $x_{i+1} = \begin{cases} 1 & x_i = 0 \\ \frac{1}{mod(x_i, 1)} & otherwise \end{cases}$ | (0,1) |
| Iterative [35] | $x_{i+1} = \sin\left(\frac{a\pi}{x_i}\right), a = 0.7$ | (-1,1) |
| Logistic [36],[37] | $x_{i+1} = ax_i(1 - x_i), a = 4$ | (0,1) |
| Piecewise [38] | $x_{i+1} = \begin{cases} \frac{x_i}{p} & 0 \leq x_i < p \\ \frac{x_i - p}{0.5 - p} & p \leq x_i < 0.5 \\ \frac{1 - p - x_i}{0.5 - p} & 0.5 \leq x_i < 1 - p \\ \frac{1 - x_i}{p} & 1 - p \leq x_i < 1 \end{cases}$ | (0,1) |
| Sine [39] | $x_{i+1} = \frac{a}{4}\sin(\pi x_i), a = 4$ | (0,1) |
| Singer [32] | $x_{i+1} = \mu(7.86x_i - 23.31x_i^2 + 28.75x_i^3 - 13.302875x_i^4), \mu = 1.07$ | (0,1) |
| Sinusoidal [40] | $x_{i+1} = ax_i^2 \sin(\pi x_i), a = 2.3$ | (0,1) |
| Tent [41] | $x_{i+1} = \begin{cases} \frac{x_i}{0.7} & x_i < 0.7 \\ \frac{10}{3}(1 - x_i) & x_i > 0.7 \end{cases}$ | (0,1) |

Table 2 Cec2019 Benchmark Functions

| No. | FUNCTIONS | $F_i^* = F_i(x^*)$ | D | Search Range |
|---|---|---|---|---|
| 1 | Storn's Chebyshev Polynomial Fitting Problem | 1 | 9 | [-8192, 8192] |
| 2 | Inverse Hilbert Matrix Problem | 1 | 16 | [-16384, 16384] |
| 3 | Lennard-Joes Minimum Energy Cluster | 1 | 18 | [-4, 4] |
| 4 | Rastrigin's Function | 1 | 10 | [-100, 100] |
| 5 | Griewangk's Function | 1 | 10 | [-100, 100] |
| 6 | Weierstrass Function | 1 | 10 | [-100, 100] |
| 7 | Modified Schwefel's Function | 1 | 10 | [-100, 100] |
| 8 | Expand Schaffer's F6 function | 1 | 10 | [-100, 100] |
| 9 | Happy Cat Function | 1 | 10 | [-100, 100] |
| 10 | Ackley Function | 1 | 10 | [-100, 100] |

Table 3 Average Results and *p*-value of Chaotic Maps F1 and F2 on CEC2019 Benchmark Functions.

| F1 | Mean | Std. | *p*-Value | F2 | Mean | Std. | *p*-Value |
|---|---|---|---|---|---|---|---|
| FDO | 5.56E+10 | 48042650116 | 0.0001 | FDO | 22.993 | 20.77176 | 3.17E-252 |
| CFDO1 | 3.57E+10 | 46995630286 | 7.62542E-05 | CFDO1 | 20.2261 | 2.833845 | 1 |
| CFDO2 | 1.63E+11 | 1.42803E+11 | 0.999998532 | CFDO2 | 18.9447 | 0.51868 | 1 |
| CFDO3 | 3.23E+10 | 78864744102 | 2.51642E-10 | CFDO3 | 18.5048 | 0.573031 | 1 |
| CFDO4 | 5.12E+10 | 93204852223 | 0.009703409 | CFDO4 | 3.83E+03 | 4739.81 | 1 |
| CFDO5 | 1.81E+10 | 30659064404 | 1.9228E-12 | CFDO5 | 18.8393 | 1.030596 | 1 |
| CFDO6 | 4.44E+10 | 73797023388 | 0.00053268 | CFDO6 | 18.3446 | 0.635342 | 1 |
| CFDO7 | 2.47E+10 | 46019833354 | 3.92458E-10 | CFDO7 | 19.1817 | 2.551178 | 1 |
| CFDO8 | 1.39E+11 | 3.20398E+11 | 0.180970458 | CFDO8 | 19.0457 | 6.687333 | 0.000174 |
| CFDO9 | 8.02E+05 | 3109104.768 | 0 | CFDO9 | 19.3878 | 0.500974 | 1 |
| CFDO10 | 3.52E+11 | 6.57231E+11 | 1 | CFDO10 | 1.52E+04 | 184.328 | 1 |

Table 4  Average Results and *p*-value of Chaotic Maps F3 and F4 on CEC2019 Benchmark Functions.

| F3 | Mean | Std. | *p*-Value | F4 | Mean | Std. | *p*-Value |
|---|---|---|---|---|---|---|---|
| FDO | 12.7024 | 5.33E-15 | 0.363 | FDO | 118.1335 | 41.52297 | 0.095 |
| CFDO1 | 12.7024 | 5.33E-15 | 1 | CFDO1 | 12132 | 5343.528 | 1 |
| CFDO2 | 12.7024 | 3.96E-05 | 1 | CFDO2 | 402.0163 | 520.7971 | 1 |
| CFDO3 | 12.7024 | 5.33E-15 | 1 | CFDO3 | 156.9835 | 98.5411 | 0.997662 |
| CFDO4 | 12.7024 | 5.33E-15 | 1 | CFDO4 | 1745.2 | 1656.898 | 1 |
| CFDO5 | 12.7024 | 5.33E-15 | 1 | CFDO5 | 114.0643 | 51.15493 | 0.095256 |
| CFDO6 | 12.7024 | 5.33E-15 | 1 | CFDO6 | 115.8564 | 55.31845 | 0.295408 |
| CFDO7 | 12.7024 | 5.33E-15 | 1 | CFDO7 | 143.4651 | 61.06715 | 0.993392 |
| CFDO8 | 12.7025 | 0.000682 | 1 | CFDO8 | 281.8257 | 347.6287 | 0.999976 |
| CFDO9 | 12.7024 | 7.92E-05 | 1 | CFDO9 | 711.5787 | 701.7925 | 1 |
| CFDO10 | 12.7089 | 0.002937 | 1 | CFDO10 | 38050 | 8813.867 | 1 |

Table 5  Average Results and *p*-value of Chaotic Maps F5 and F6 on CEC2019 Benchmark Functions.

| F5 | Mean | Std. | *p*-Value | F6 | Mean | Std. | *p*-Value |
|---|---|---|---|---|---|---|---|
| FDO | 1.332 | 0.192678 | 0.007 | FDO | 12.511 | 0.840741 | 2.196E-06 |
| CFDO1 | 2.5862 | 0.202408 | 1 | CFDO1 | 12.4389 | 0.773043 | 0.380605 |
| CFDO2 | 2.1496 | 0.30528 | 1 | CFDO2 | 12.1053 | 1.461689 | 0.116594 |
| CFDO3 | 1.8049 | 0.211747 | 1 | CFDO3 | 11.8337 | 1.738952 | 0.035933 |
| CFDO4 | 1.6377 | 0.446271 | 1 | CFDO4 | 10.9224 | 2.265566 | 9.44E-06 |
| CFDO5 | 1.489 | 0.330753 | 0.999841 | CFDO5 | 10.8164 | 2.232193 | 1.16E-08 |
| CFDO6 | 1.5739 | 0.311248 | 1 | CFDO6 | 10.5732 | 2.866736 | 1.94E-05 |
| CFDO7 | 1.5284 | 0.369671 | 0.999995 | CFDO7 | 11.3231 | 1.956137 | 6.94E-05 |
| CFDO8 | 1.7595 | 0.739106 | 0.999262 | CFDO8 | 10.443 | 1.990047 | 1.17E-12 |
| CFDO9 | 1.8357 | 0.662118 | 1 | CFDO9 | 11.6539 | 1.776806 | 4.90E-04 |
| CFDO10 | 7.6212 | 0.518032 | 1 | CFDO10 | 13.5967 | 0.868955 | 1 |

Cite as : Mohammed, H.M., Rashid, T.A. Chaotic fitness-dependent optimizer for planning and engineering design. Soft Comput (2021). https://doi.org/10.1007/s00500-021-06135-z

Table 6 Average Results and *p*-value of Chaotic Maps F7 and F8 on CEC2019 Benchmark Functions.

| F7 | Mean | Std. | *p*-Value | F8 | Mean | Std. | *p*-Value |
|---|---|---|---|---|---|---|---|
| FDO | 717.0504 | 261.7898 | 1.0424E-06 | FDO | 6.2057 | 0.526052 | 0.131 |
| CFDO1 | 1026 | 287.4806 | 1 | CFDO1 | 6.6992 | 0.463863 | 1 |
| CFDO2 | 1028.6 | 494.8696 | 0.999987 | CFDO2 | 6.7221 | 0.56229 | 1 |
| CFDO3 | 649.3674 | 354.5131 | 0.080372 | CFDO3 | 6.3994 | 0.5816 | 0.969265 |
| CFDO4 | 744.0466 | 352.0682 | 0.481355 | CFDO4 | 6.09 | 0.611987 | 0.076941 |
| CFDO5 | 645.103 | 391.1508 | 0.046179 | CFDO5 | 6.1785 | 0.515245 | 0.320062 |
| CFDO6 | 598.8501 | 391.585 | 0.001874 | CFDO6 | 6.3724 | 0.578174 | 0.982305 |
| CFDO7 | 718.8622 | 361.5221 | 0.328468 | CFDO7 | 6.1001 | 0.675956 | 0.271666 |
| CFDO8 | 643.2904 | 381.5608 | 0.005417 | CFDO8 | 6.0375 | 0.483469 | 0.022197 |
| CFDO9 | 869.4566 | 522.3861 | 0.858881 | CFDO9 | 6.5703 | 0.690458 | 0.999988 |
| CFDO10 | 2011.2 | 426.62 | 1 | CFDO10 | 7.9354 | 0.400813 | 1 |

Table 7 Average Results and *p*-value of Chaotic Maps F9 and F10 on CEC2019 Benchmark Functions.

| F9 | Mean | Std. | *p*-Value | F10 | Mean | Std. | *p*-Value |
|---|---|---|---|---|---|---|---|
| FDO | 4.5682 | 0.756012 | 3.7992E-43 | FDO | 20.4023 | 0.199214 | 6.397E-131 |
| CFDO1 | 640.5673 | 212.7795 | 1 | CFDO1 | 20.4772 | 0.178562 | 0.987588 |
| CFDO2 | 12.9474 | 25.58764 | 1 | CFDO2 | 20.4381 | 0.259686 | 0.892316 |
| CFDO3 | 4.95 | 0.946272 | 0.990887 | CFDO3 | 20.132 | 0.183273 | 5.28E-14 |
| CFDO4 | 148.1725 | 220.5425 | 1 | CFDO4 | 19.9684 | 0.727757 | 0 |
| CFDO5 | 4.2436 | 0.684265 | 0.004127 | CFDO5 | 20.0974 | 0.155022 | 0 |
| CFDO6 | 4.1758 | 0.762961 | 0.000681 | CFDO6 | 20.0963 | 0.167575 | 0 |
| CFDO7 | 4.2879 | 1.261513 | 0.000121 | CFDO7 | 19.9609 | 0.899888 | 1.05E-14 |
| CFDO8 | 4.2719 | 0.606944 | 0.005417 | CFDO8 | 20.0081 | 0.021456 | 0 |
| CFDO9 | 5.2566 | 1.339834 | 0.999792 | CFDO9 | 20.0622 | 0.116389 | 0 |
| CFDO10 | 2570.8 | 1320.496 | 1 | CFDO10 | 20.949 | 0.235344 | 1 |

Table 8  Comparison Results of WOA, FDO, GWO, and CFDO on CEC2019.

| F | CSO | | GA | | PSO | | CFDO | | FDO | |
|---|---|---|---|---|---|---|---|---|---|---|
| | Avg. | STD | Avg. | STD | Avg. | STD | Avg. | Std. | Avg. | Std. |
| 1 | 3.66E+09 | 3.55E+09 | **5.32E+04** | 7.04E+04 | 1.47127E + 12 | 1.32362E + 12 | 1.39E+11 | 3.20E+11 | 5.56E+10 | 4.80E+10 |
| 2 | 19.53886 | 0.608508 | **17.3502** | 17.3491 | 1.52E+04 | 3.73E+03 | 1.90E+01 | 6.69E+00 | 2.30E+01 | 2.08E+01 |
| 3 | 13.70241 | 8.33E-06 | **12.7024** | 13.7024 | **1.27E+01** | 9.03E-15 | 1.27E+01 | 6.82E-04 | **1.27E+01** | 5.33E-15 |
| 4 | 198.9105 | 81.32489 | 6.23E+04 | 61986.61 | **1.68E+01** | 8.20E+00 | 2.82E+02 | 3.48E+02 | 1.18E+02 | 4.15E+01 |
| 5 | 2.753796 | 0.192018 | 7.5396 | 7.2765 | **1.14E+00** | 8.94E-02 | 1.76E+00 | 7.39E-01 | 1.33E+00 | 1.93E-01 |
| 6 | 11.66279 | 0.732559 | **7.4005** | 6.6877 | 9.31E+00 | 1.69E + 00 | 1.04E+01 | 1.99E+00 | 1.25E+01 | 8.41E-01 |
| 7 | 457.0046 | 141.4665 | 791.742 | 697.8964 | **1.61E+02** | 1.04E+02 | 6.43E+02 | 3.82E+02 | 7.17E+02 | 2.62E+02 |
| 8 | 5.679993 | 0.47298 | 6.1004 | 5.8228 | **5.22E+00** | 7.87E-01 | 6.04E+00 | 4.83E-01 | 6.21E+00 | 5.26E-01 |
| 9 | 15.06303 | 11.49835 | 5.31E+03 | 5.29E+03 | **2.37E+00** | 1.84E-02 | 4.27E+00 | 6.07E-01 | 4.57E+00 | 7.56E-01 |
| 10 | 21.40961 | 0.087453 | 20.1059 | 20.0236 | 2.03E+01 | 1.29E-01 | **2.00E+01** | 2.15E-02 | 2.04E+01 | 1.99E-01 |

Table 9 CEC2013 Benchmark Functions, Types, and Ranges.

| | No. | Functions | $f_i^*=f_i(x^*)$ |
|---|---|---|---|
| **Unimodal Functions** | 1 | Sphere Function | -1400 |
| | 2 | Rotated High Conditioned Elliptic Function | -1300 |
| | 3 | Rotated Bent Cigar Function | -1200 |
| | 4 | Rotated Discus Function | -1100 |
| | 5 | Different Powers Function | -1000 |
| **Basic Multimodal Functions** | 6 | Rotated Rosenbrock's Function | -900 |
| | 7 | Rotated Schaffers F7 Function | -800 |
| | 8 | Rotated Ackley's Function | -700 |
| | 9 | Rotated Weierstrass Function | -600 |
| | 10 | Rotated Griewank's Function | -500 |
| | 11 | Rastrigin's Function | -400 |
| | 12 | Rotated Rastrigin's Function | -300 |
| | 13 | Non-Continuous Rotated Rastrigin's Function | -200 |
| | 14 | Schwefel's Function | -100 |
| | 15 | Rotated Schwefel's Function | 100 |
| | 16 | Rotated Katsuura Function | 200 |
| | 17 | Lunacek Bi_Rastrigin Function | 300 |
| | 18 | Rotated Lunacek Bi_Rastrigin Function | 400 |
| | 19 | Expanded Griewank's plus Rosenbrock's Function | 500 |
| | 20 | Expanded Scaffer's F6 Function | 600 |
| **Composition Functions** | 21 | Composition Function 1 (n=5,Rotated) | 700 |
| | 22 | Composition Function 2 (n=3,Unrotated) | 800 |
| | 23 | Composition Function 3 (n=3,Rotated) | 900 |
| | 24 | Composition Function 4 (n=3,Rotated) | 1000 |
| | 25 | Composition Function 5 (n=3,Rotated) | 1100 |
| | 26 | Composition Function 6 (n=5,Rotated) | 1200 |
| | 27 | Composition Function 7 (n=5,Rotated) | 1300 |
| | 28 | Composition Function 8 (n=5,Rotated) | 1400 |
| Search Range: $[-100,100]^D$ | | | |

Table 10 Comparison Results of CFDO against Five Algorithms using CEC2013.

| F. | CJADE | | IPOP-CMA-ES | | NBIPOP-CMA-ES | | SHADE | | GSA | | CFDO | |
|---|---|---|---|---|---|---|---|---|---|---|---|---|
| | Avg. | STD | Avg. | Std | Avg. | STD | Avg. | STD | Avg. | STD | Avg. | STD |
| 1 | -1400 | 5.97E-14 | 1.00E-08 | 0 | 0 | 0 | 0 | 0 | −1.40E+03 | 0.00E+00 | 7.01E+04 | 95682.65 |
| 2 | 3045.822 | 2285.002 | 1.00E-08 | 0 | **0** | 0 | 9000 | -7470 | 7.32E+06 | 1.39E+06 | 3.63E+09 | 7.8E+09 |
| 3 | 2944296 | 7496535 | 1.00E-08 | 0 | **0** | 0 | 40.2 | -213 | 5.33E+09 | 2.35E+09 | 1.97E+22 | 1.35E+22 |
| 4 | 3379.984 | 11721.86 | 1.00E-08 | 0 | 0 | 0 | 0.000192 | 0.000301 | 6.70E+04 | 3.50E+03 | 7.11E+04 | 3728.671 |
| 5 | -1000 | 0 | 1.00E-08 | 0 | 0 | 0 | 0 | 0 | −1.00E+03 | 6.20E-13 | 7.92E+04 | 5735.703 |
| 6 | **-895.466** | 9.975666 | 1.00E-08 | 0 | 0 | 0 | 0.596 | -3.73 | −8.33E+02 | 1.36E+01 | 2.16E+04 | 1783.143 |
| 7 | **-779.176** | 13.19676 | 7.01E-02 | 1.56E-01 | 2.313 | 6.049 | 4.6 | -5.39 | −7.33E+02 | 1.20E+01 | 1.23E+08 | 58423514 |
| 8 | **-679.064** | 0.073209 | 20.9 | 6.23E-02 | 20.942 | 0.048 | 20.7 | -0.176 | -6.79E+02 | 4.61E-02 | **-678.649** | 0.088866 |
| 9 | **-571.738** | 1.633441 | 4.34 | 1.72 | 3.3 | 1.383 | 27.5 | -1.77 | -5.67E+02 | 2.64E+00 | -5.44E+02 | 1.571278 |
| 10 | -499.953 | 0.026126 | 1.00E-08 | 0 | 0 | 0 | 0.0769 | -0.0358 | **−5.00E+02** | 4.78E-02 | 1.27E+04 | 1353.254 |
| 11 | **-400** | 0 | 2.25 | 1.05 | 3.043 | 1.413 | 0 | 0 | -1.05E+02 | 2.01E+01 | 783.2169 | 37.66595 |
| 12 | **-274.825** | 6.257304 | 1.72 | 1.23 | 2.907 | 1.376 | 23 | -3.73 | 3.12E+01 | 2.58E+01 | 829.4422 | 35.16362 |
| 13 | **-99.8952** | 0.03807 | 2.16 | 1.3 | 2.778 | 1.453 | 50.3 | -13.4 | 2.65E+02 | 3.89E+01 | 926.2928 | 37.63335 |
| 14 | 3192.957 | 326.1168 | 708 | 294 | 810.125 | 360.294 | **0.0318** | -0.0233 | 3.91E+03 | 5.71E+02 | 1.12E+04 | 411.0682 |
| 15 | **201.3478** | 0.70741 | 259 | 118 | 765.493 | 294.867 | 3220 | -264 | 3.64E+03 | 5.08E+02 | 9.84E+03 | 318.5119 |
| 16 | 330.4337 | 1.23E-06 | **3.75E-01** | 2.65E-01 | 0.44 | 0.926 | 0.913 | -0.185 | 2.00E+02 | 3.41E-03 | 206.0202 | 1.276843 |
| 17 | 464.117 | 5.633772 | 34.3 | 1.86 | 34.419 | 1.869 | **30.4** | -3.8E-14 | 3.66E+02 | 8.21E+00 | 1.34E+03 | 17.94635 |
| 18 | 501.1961 | 0.183779 | **40.1** | 18.7 | 62.289 | 45.591 | 72.5 | -5.58 | 4.56E+02 | 5.02E+00 | 1.42E+03 | 20.36665 |
| 19 | 610.7729 | 0.683516 | 2.24 | 5.66E-01 | 2.228 | 0.341 | **1.36** | -0.12 | 5.10E+02 | 2.63E+00 | 1.36E+06 | 265853 |
| 20 | 1015.375 | 73.46999 | 14.4 | 7.38E-01 | 12.94 | 0.598 | **10.5** | -0.604 | 6.15E+02 | 1.68E-01 | 615 | 0 |
| 21 | 1015.375 | 73.46999 | **188** | 32.5 | 192.157 | 27.152 | 309 | -56.5 | 1.01E+03 | 3.64E+01 | 3.38E+03 | 27.38683 |
| 22 | 894.611 | 31.41581 | 533 | 363 | 838.392 | 459.988 | **98.1** | -25.2 | 7.22E+03 | 3.64E+01 | 1.21E+04 | 482.8466 |
| 23 | 4181.815 | 347.4164 | **269** | 141 | 667.086 | 289.554 | 3510 | -411 | 6.79E+03 | 3.99E+02 | 1.16E+04 | 318.5163 |
| 24 | 1238.289 | 19.97989 | 200 | 6.16E-04 | **161.757** | 30.045 | 205 | -5.29 | 1.32E+03 | 5.82E+01 | 1937.191 | 63.61232 |
| 25 | 1374.26 | 12.92483 | 240 | 5.12 | **219.984** | 11.094 | 259 | -19.6 | 1.49E+03 | 1.31E+01 | 1.60E+03 | 47.80801 |
| 26 | 1441.275 | 65.42218 | 216 | 36.7 | **158.223** | 29.999 | 202 | -14.8 | 1.56E+03 | 1.97E+01 | 2.44E+03 | 584.1771 |
| 27 | 2129.396 | 182.8821 | **300** | 9.34E-03 | 468.925 | 73.77 | 388 | -109 | 2.23E+03 | 9.49E+01 | 4.32E+03 | 250.2937 |
| 28 | 1735.335 | 193.5383 | **245** | 90.1 | 268.627 | 73.458 | 300 | 0 | 5.07E+03 | 2.41E+02 | 1.00E+04 | 309.7208 |

**Table 11 CEC2005 Benchmark Functions, Types, and Ranges.**

| No. | Functions | $F_1(x^*) = f\_bias_i$ | D | Search Range |
|---|---|---|---|---|
| **Unimodal Benchmark Functions** | | | | |
| 1 | Shifted Sphere Function | -450 | 10,30,50 | [-100, 100] |
| 2 | Shifted Schwefel's Problem 1.2 | -450 | 10,30,50 | [-100, 100] |
| 3 | Shifted Rotated High Conditioned Elliptic Function | -450 | 10,30,50 | [-100, 100] |
| 4 | Shifted Schwefel's Problem 1.2 with Noise in Fitness | -450 | 10,30,50 | [-100, 100] |
| 5 | Schwefel's Problem 2.6 with Global Optimum on Bounds | -310 | 10,30,50 | [-100, 100] |
| **Multimodal Benchmark functions** | | | | |
| 6 | Shifted Rosenbrock's Function | 390 | 10,30,50 | [-100, 100] |
| 7 | Shifted Rotated Griewank's Function without Bounds | -180 | 10,30,50 | [0, 600] |
| 8 | Shifted Rotated Ackley's Function with Global Optimum on Bounds | -140 | 10,30,50 | [-32, 32] |
| 9 | Shifted Rastrigin's Function | -330 | 10,30,50 | [-5, 5] |
| 10 | Shifted Rotated Rastrigin's Function | -330 | 10,30,50 | [-5, 5] |
| 11 | Shifted Rotated Weierstrass Function | 90 | 10,30,50 | [-0.5, 0.5] |
| 12 | Schwefel's Problem 2.13 | -460 | 10,30,50 | [-π, π] |
| **Expanded Benchmark functions** | | | | |
| 13 | Expanded Extended Griewank's plus Rosenbrock's Function (F8F2) | -130 | 10,30,50 | [-3, 1] |
| 14 | Shifted Rotated Expanded Scaffer's F6 | -300 | 10,30,50 | [-100, 100] |
| **Hybrid Composition Functions** | | | | |
| 15 | Hybrid Composition Function | 120 | 10,30,50 | [-5, 5] |
| 16 | Rotated Hybrid Composition Function | 120 | 10,30,50 | [-5, 5] |
| 17 | Rotated Hybrid Composition Function with Noise in Fitness | 120 | 10,30,50 | [-5, 5] |
| 18 | Rotated Hybrid Composition Function | 10 | 10,30,50 | [-5, 5] |
| 19 | Rotated Hybrid Composition Function with a Narrow Basin for the Global Optimum | 10 | 10,30,50 | [-5, 5] |
| 20 | Rotated Hybrid Composition Function with the Global Optimum on the Bounds | 10 | 10,30,50 | [-5, 5] |
| 21 | Rotated Hybrid Composition Function | 360 | 10,30,50 | [-5, 5] |
| 22 | Rotated Hybrid Composition Function with High Condition Number Matrix | 360 | 10,30,50 | [-5, 5] |
| 23 | Non-Continuous Rotated Hybrid Composition Function | 360 | 10,30,50 | [-5, 5] |
| 24 | Rotated Hybrid Composition Function | 260 | 10,30,50 | [-5, 5] |
| 25 | Rotated Hybrid Composition Function without Bounds | 260 | 10,30,50 | [2, 5] |

Table 12 Comparison Results of CFDO against BSO and CBSO using CEC2005.

| F. | BSO Avg. | BSO STD | CBSO Avg. | CBSO STD | CFDO Avg. | CFDO STD |
|---|---|---|---|---|---|---|
| 1 | -4.50E+02 | 3.50E−14 | -4.50E+02 | 3.17E−14 | 6.37E-28 | -4.50E+02 |
| 2 | -4.48E+02 | 9.36E−01 | -4.48E+02 | 9.91E−01 | 1.31E-24 | -4.48E+02 |
| 3 | 2.04E+06 | 7.23E+05 | 1.78E+06 | 6.82E+05 | **1.23E+06** | 1.23E+06 |
| 4 | 2.78E+04 | 8.05E+03 | 2.05E+04 | 6.24E+03 | **1.06E+04** | 1.06E+04 |
| 5 | 4.70E+03 | 1.22E+03 | 4.15E+03 | 7.56E+02 | **6.95E-10** | 6.95E-10 |
| 6 | 1.26E+03 | 9.48E+02 | 9.48E+02 | 3.73E+02 | **7.64E-21** | 7.64E-21 |
| 7 | 6.25E+03 | 3.25E+02 | 6.05E+03 | 3.28E+02 | **1.27E+03** | 1.27E+03 |
| 8 | **-120** | 9.90E−02 | **-120** | 6.22E−02 | 2.02E+01 | -1.20E+02 |
| 9 | **-286** | 1.27E+01 | **-286** | 9.58E+00 | 1.19E+01 | -2.86E+02 |
| 10 | **-2.91E+02** | 8.79E+00 | **-2.91E+02** | 1.05E+01 | 6.27E+01 | -2.91E+02 |
| 11 | 1.10E+02 | 2.51E+00 | 1.09E+02 | 2.70E+00 | **4.65E+00** | 4.65E+00 |
| 12 | 2.84E+04 | 1.99E+04 | 2.43E+04 | 1.68E+04 | **2.12E+02** | 2.12E+02 |
| 13 | **-1.26E+02** | 1.05E+00 | **-1.26E+02** | 9.25E−01 | 1.98E+00 | -1.26E+02 |
| 14 | **-2.87E+02** | 3.78E−01 | **-2.87E+02** | 3.61E−01 | 4.0686 | -2.87E+02 |
| 15 | 5.43E+02 | 7.94E+01 | 5.15E+02 | 6.63E+01 | | |
| 16 | 2.87E+02 | 1.34E+02 | 2.63E+02 | 1.42E+02 | | |
| 17 | 3.10E+02 | 1.57E+02 | 2.87E+02 | 1.30E+02 | | |
| 18 | 9.17E+02 | 1.36E+00 | 9.16E+02 | 1.20E+00 | | |
| 19 | 9.16E+02 | 1.07E+00 | 9.16E+02 | 1.28E+00 | | |
| 20 | 9.16E+02 | 1.36E+00 | 9.16E+02 | 1.17E+00 | | |
| 21 | 9.27E+02 | 1.37E+02 | 8.87E+02 | 1.01E+02 | | |
| 22 | 1.21E+03 | 1.99E+01 | 1.22E+03 | 1.76E+01 | | |
| 23 | 9.48E+02 | 1.38E+02 | 8.95E+02 | 1.40E+00 | | |
| 24 | 4.67E+02 | 6.23E+00 | 4.60E+02 | 1.67E−12 | | |
| 25 | 1.88E+03 | 4.44E+00 | 1.88E+03 | 3.39E+00 | | |

Table 13 Comparison of WOA, GWO, CGWO, CFDO, and FDO for Pressure Vessel Design

| Algorithm | Avg. | Std. |
|---|---|---|
| WOA | 1.36E+04 | 1.27E+04 |
| GWO | 6.16E+03 | 379.674 |
| CGWO | 5.78E+03 | 254.505 |
| FDO | 5.33E+04 | 47583.22 |
| CFDO | **5.14E+03** | 6736.938 |

Table 14 Task Assignment Problem.

| Tasks / Employees | Employee 1 | Employee 2 | Employee 3 | Employee 4 | Employee 5 |
|---|---|---|---|---|---|
| Task 1 | 45 | 65 | 34 | 45 | 76 |
| Task 2 | 83 | 23 | 44 | 82 | 32 |
| Task 3 | 22 | 67 | 64 | 45 | 12 |
| Task 4 | 76 | 76 | 26 | 32 | 65 |
| Task 5 | 10 | 43 | 19 | 43 | 99 |

Table 9 Comparison of CFDO and FDO for Solving Task Assignment Problem.

| Algorithm | Avg. | Std. |
|---|---|---|
| CFDO | **50** | **0** |
| FDO | 260.6623 | 59.40106 |